\documentclass{article}

\usepackage{times}
\usepackage{graphicx} 
\usepackage{subfigure} 

\usepackage{natbib}

\usepackage{hyperref}

\usepackage[accepted]{icml2015}

\usepackage[T1]{fontenc}
\usepackage[utf8x]{inputenc}
\usepackage{algpseudocode}
\usepackage{algorithm}
\usepackage{amsmath}
\usepackage{amssymb}
\usepackage{amsfonts}
\usepackage[outdir=./]{epstopdf}
\usepackage{afterpage}

\newcommand{\Xcal}{\mathcal{X}}
\newcommand{\Acal}{\mathcal{A}}

\newcommand{\x}{\mathbf{x}}
\newcommand{\Rbb}{\mathbb{R}}
\newcommand{\W}{\mathbf{W}}
\renewcommand{\b}{\mathbf{b}}
\renewcommand{\v}{\mathbf{v}}
\newcommand{\h}{\mathbf{h}}
\newcommand{\y}{\mathbf{y}}
\newcommand{\Scal}{\mathcal{S}}
\newcommand{\Ocal}{\mathcal{O}}

\icmltitlerunning{Multiple Locally Linear Kernel Machines}

\begin{document} 

\twocolumn[
\icmltitle{Multiple Locally Linear Kernel Machines}

\icmlauthor{David Picard}{david.picard@enpc.fr}
\icmladdress{LIGM, Ecole des Ponts, Univ Gustave Eiffel, CNRS, Marne-la-Vallée, France}

\icmlkeywords{boring formatting information, machine learning, ICML}

\vskip 0.3in
]

\begin{abstract}
In this paper we propose a new non-linear classifier based on a combination of locally linear classifiers.
A well known optimization formulation is given as we cast the problem in a $\ell_1$ Multiple Kernel Learning (MKL) problem using many locally linear kernels.
Since the number of such kernels is huge, we provide a scalable generic MKL training algorithm handling streaming kernels.
With respect to the inference time, the resulting classifier fits the gap between high accuracy but slow non-linear classifiers (such as classical MKL) and fast but low accuracy linear classifiers.
\end{abstract}

\section{Introduction}

SVMs were shown to provide very accurate classifiers, and are consequently used in a growing number of applications.
SVM training algorithms and subsequent inference efficiency can be separated in two groups, depending on the kernel being linear or not.
When dealing with linear SVMs, very efficient training algorithms based on stochastic gradient descent such as~\cite{shalevshwartz07icml,bordes09jmlr,leroux12nips} allow to deal with large datasets in a short amount of time.
The inference time of linear SVMs is clearly their main advantage as it can efficiently be optimized for modern computer architecture.
However, many classification problems are not linearly separable and consequently a non linear kernel is required.

Kernel SVMs also benefit from recent developments in stochastic coordinate methods such as~\cite{bordes05jmlr, shalevshwartz13jmlr, shalevshwartz14icml} and come now with efficient training procedures.
The main drawback of Kernel SVMs is the inference time which is proportional to the number of support vectors.
This number has been experimentally shown to grow linearly with the size of the training set~\cite{bordes05jmlr}.
Most of the speed-up is obtained by limiting the number of support vector~\cite{dekel06nips}.
For example, in~\cite{ertekin11pami}, the authors proposed to ignore outliers to reduce the number of support vectors, but the problem becomes non-convex and the resulting classifier still has a high inference time.

With respect to the inference cost, a much more efficient alternative is the use of almost linear classifiers such as Locally Linear SVM (LLSVM) proposed in~\cite{ladicky11icml}.
In LLSVM, a generative manifold learning algorithm produces a partition the input space.
An almost piece-wise linear classifier concatenating linear classifier of the different parts is then trained.
The inference cost of LLSVM is almost as fast as for linear SVMs, depending only on the number of anchor points used in the manifold learning.
However, LLSVMs are dependent on the success of the manifold learning algorithm that provides the anchor points.
Moreover, the parameter tuning of such algorithms, mainly the number of anchor points and the coding function, is often costly and difficult to perform.

In this paper, we propose a new locally linear classifier similar to LLSVM, but without the burden of the manifold learning part while keeping very efficient inference procedure.
We first define a family of locally linear kernels which are related to conformal kernels~\cite{amari99nn}.
We then used the Multiple Kernel Learning (MKL) framework~\cite{bach04icml} to select a subset of the locally linear kernels.
The $\ell_1$-norm constraint on the kernel combination leads to a limited number of selected kernels and consequently leads to low inference cost.
Our contributions can be summed up as the following: 1) we propose a new learning problem named Multiple Locally Linear Kernel Machine (MLLKM) to obtain locally linear classifiers more easily. 
2) Since MLLKM is similar to $\ell_1$-MKL, we propose a new $\ell_1$-MKL solver that can handle the high number of kernels in MLLKM.

The remaining of this paper is organized as follows.
In the next section, we review existing works on locally linear classifiers.
In Section~\ref{sec:llk}, we detail the concept of locally linear kernels.
Then, we present our proposed MLLKM problem and show it is equivalent to solving $\ell_1$-MKL.
In the same section, we present a fast algorithm to solve $\ell_1$-MKL problems on a budget and show several strategies to automatically tune the parameters and drastically lower the number of selected kernels. 
We present experiments in Section~\ref{sec:exp} before we conclude.

\section{Locally Linear Classifiers}

Generally, a locality criterion in machine learning refers to an adaptation of the model with respect to the localization in the input space.
For instance, the idea of locally linear classification is to consider a combination of different linear predictors depending on some locality criterion performed on the evaluated sample.
Remark that although the predictors are linear, the combination might not, and consequently the resulting classifier can perform non-linear separation.

In Locally Linear SVM~\cite{ladicky11icml}, a dictionary learning algorithm is used to provide a set of anchor points $\{\x_c\}$ that describe the manifold of the data.
This set is usually obtained using a k-means clustering of a large set of (unlabeled) samples, although more sophisticated method can be used.
Any sample $\x$ can then be coded by the anchor points $\x_c$ using local coordinate $\gamma_{\x_c}(\x)$ such that $\sum_{\x_c} \gamma_{\x_c}(\x) = 1$ and $\x \approx \sum_{\x_c} \gamma_{\x_c}(\x) \x_c$~\cite{yu09nips}. 
Examples of such local coordinate coding include Kernel Codebook~\cite{gemert08eccv} and Locality constraint Linear Coding~\cite{wang10cvpr}.
These local coordinate are then used to perform the combination of local classifiers:
\begin{align*}
	f(\x) = \gamma_{\x_c}(\x)^\top \W \x + \gamma_{\x_c}(\x)^\top \b,
\end{align*}
with $\W$ being the matrix which lines are the local hyperplanes and $\b$ is a vector of local biases.
To train $\W$ and $\b$, very efficient stochastic gradient descent algorithms can be used on each local hyperplane weighted by the local coordinate. 
Extension to multiclass problems can be achieve by replacing the hinge loss with a multiclass loss as presented in~\cite{fornoni13acml}.

In~\cite{gonen08icml}, the authors proposed to tackle the localization as a multiple kernel learning problem, using quasi-conformal transformation:
\begin{align*}
	K_\eta(\x_i, \x_j) = \sum_m \eta_m(\x_i)\eta_m(\x_j)K_m(\x_i, \x_j),
\end{align*}
where $\eta_m(\cdot)$ is a locality function that aims at selecting the right kernel among the $K_m$ depending on $\x_i$ and $\x_j$.
As in LLSVM, the authors proposed each locality function $\eta_m$ to be associated with an anchor point $\v_m$:

\begin{align*}
	\eta_m(\x) = \frac{exp(\langle \v_m, \x \rangle + v_{m0})}{\sum_k exp(\langle \v_k, \x \rangle + v_{ik0})}
\end{align*}

The resulting MKL problem has then 2 sets of variables, namely the dual SVM variables $\alpha$ associated with the training samples, and the anchor points $\v_m$ that define the kernels in the combination.
The LMKL algorithm consists in an alternate optimization scheme between $\alpha$ and $\v_m$, where the $\v_m$ are optimized using a gradient descent strategy.
In case of linear kernels, it is easy to see LLSVM and LMKL lead to the same family of classifiers, with LMKL allowing to tune the anchor points to the specific classification task.
However, since the LMKL objective function is non-convex, a local optimum is found.

Both methods rely on a manifold learning procedure to determine the anchor points.
The main problem of such method is that it often leads to a non-convex problem for which no guarantee on the solution can be given, and that is likely to have a high computational cost.
Moreover, the parameter tuning of such solution (\textit{i.e.}, the number of anchor points, the coding functions) is very difficult to perform. In practice, it relies on an exploration of the parameters space using cross-validation which is very costly.

In this paper, we thus propose to remove the burden of the manifold learning part by casting the selection of the anchor points and the parameters in a MKL problem.
To be able to do that, we first define a family of locally linear kernels in the next section.

\section{Locally linear kernels}
\label{sec:llk}

\begin{figure*}[!t]
	    \centering
        \subfigure[Original space $\Rbb^2$\label{fig:ori}]{
				\includegraphics[width=0.3\textwidth]{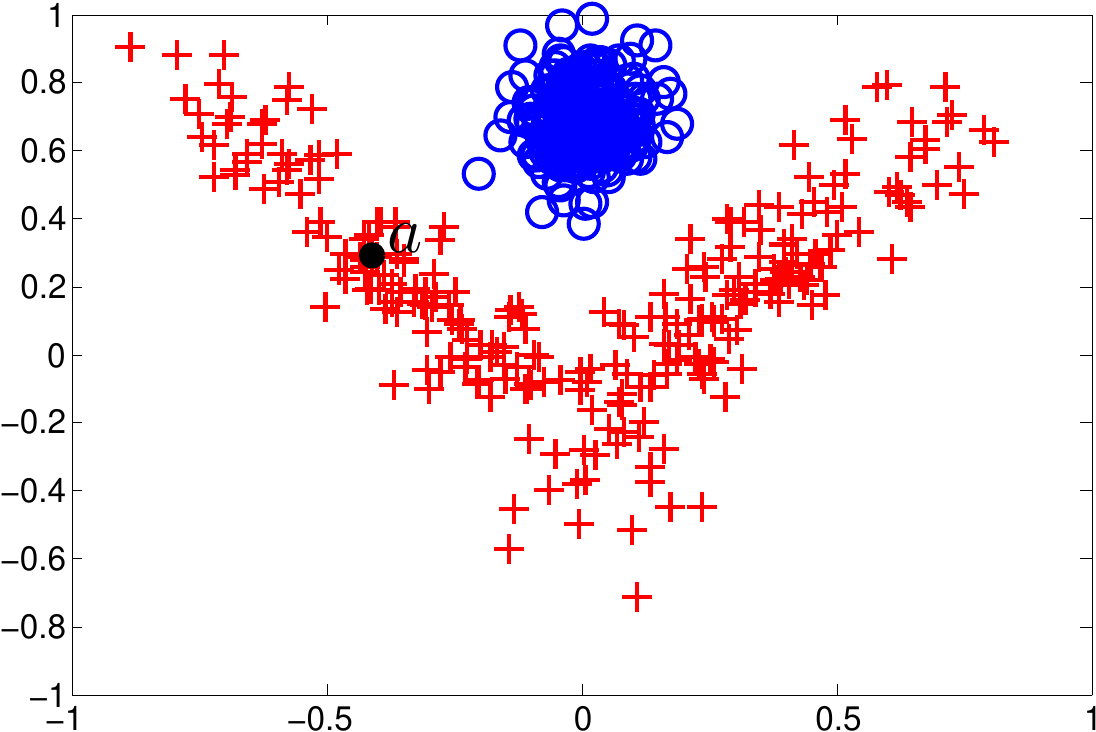}
        }
        \subfigure[Gaussian map\label{fig:mapa}]{
				\includegraphics[width=0.3\textwidth]{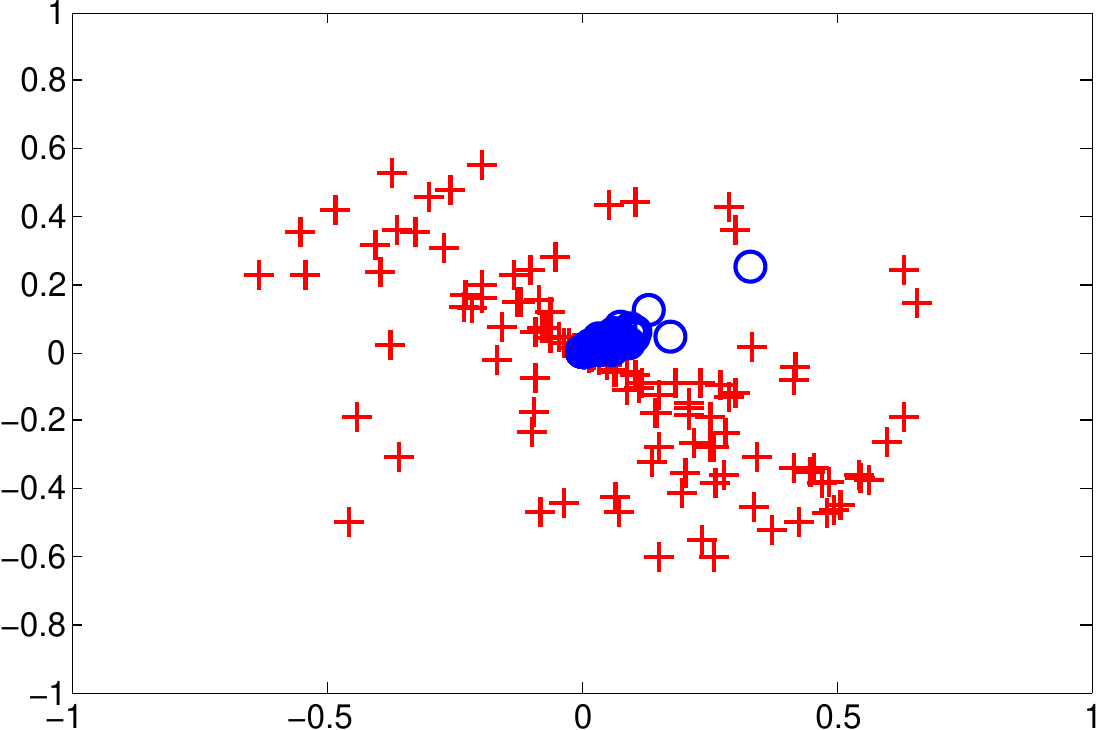}
		}
        \subfigure[Exponential map]{
                \includegraphics[width=0.3\textwidth]{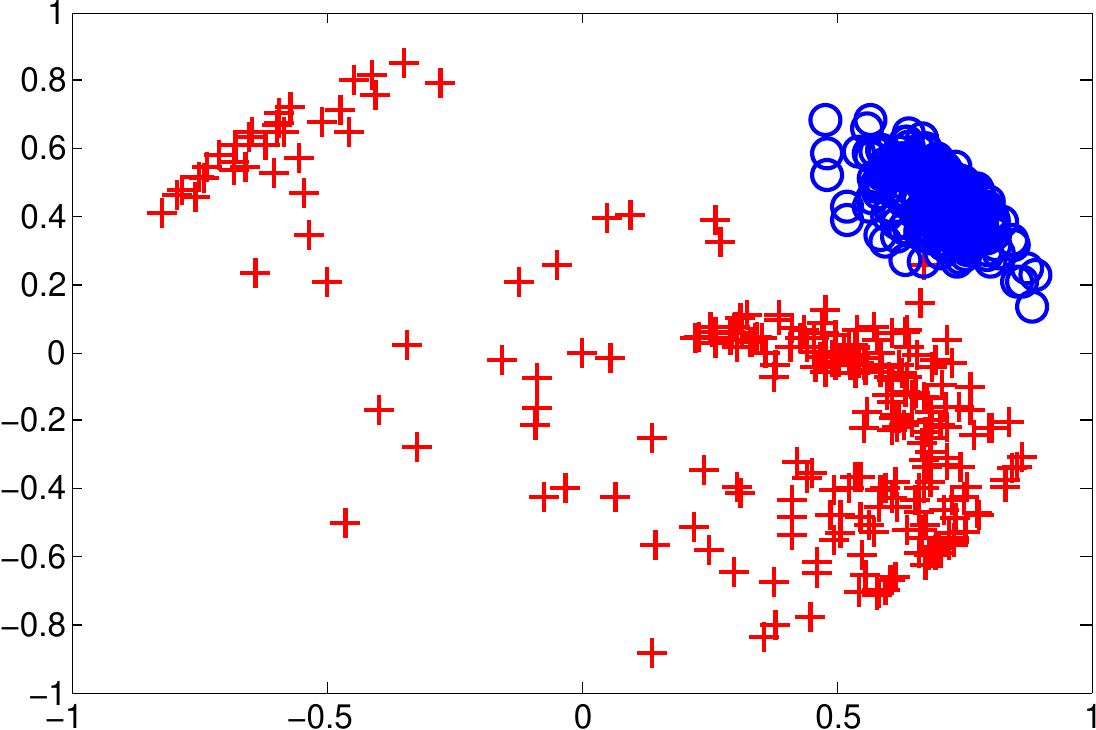}
                \label{fig:mapb}
        }
        \subfigure[Linear map]{
                \includegraphics[width=0.3\textwidth]{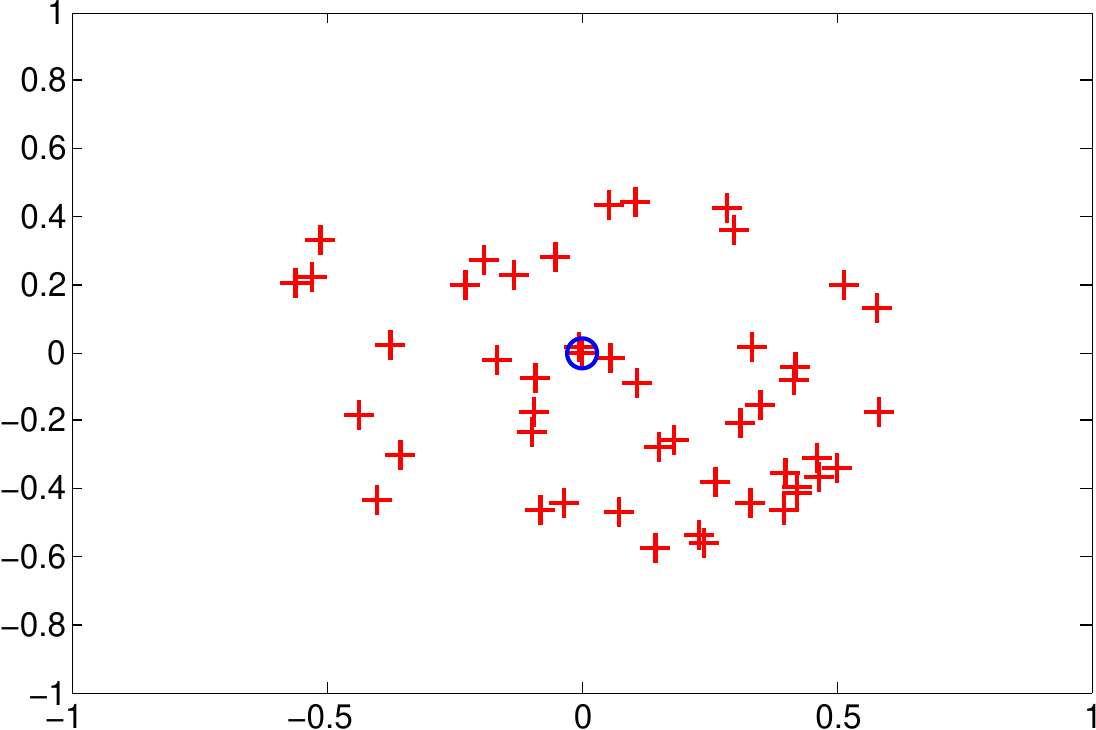}
                \label{fig:mapc}
        }
        \subfigure[Squared map]{
                \includegraphics[width=0.3\textwidth]{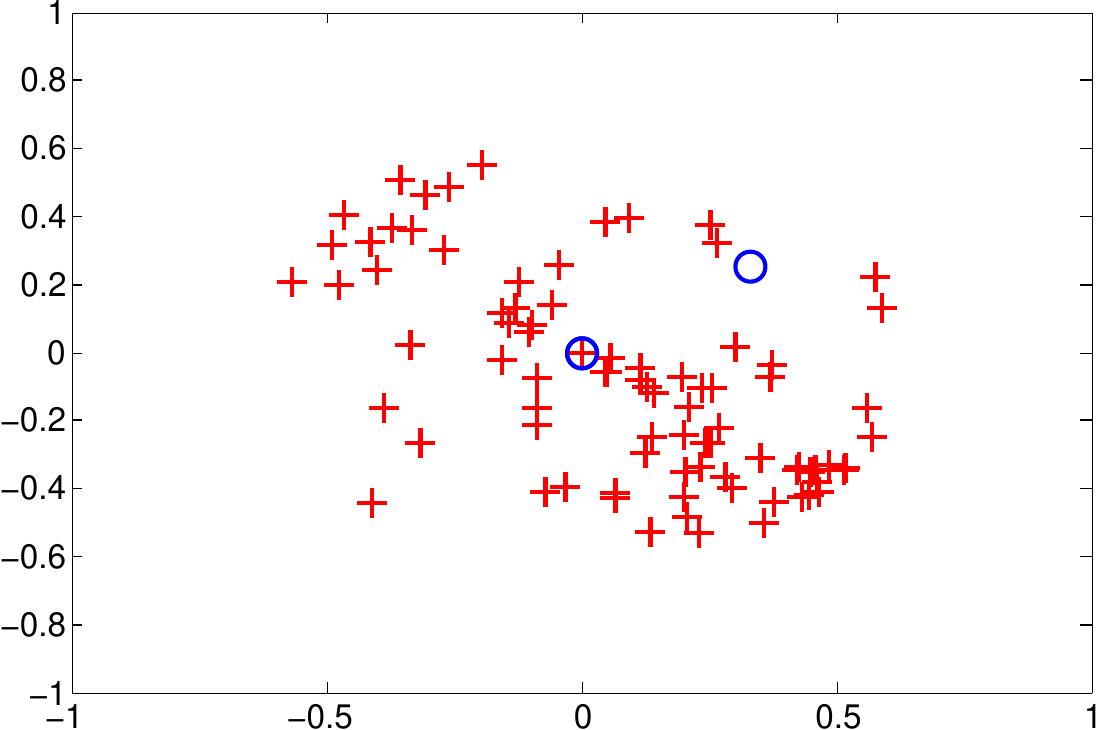}
                \label{fig:mapc}
        }
        \subfigure[Component wise Gaussian map]{
                \includegraphics[width=0.3\textwidth]{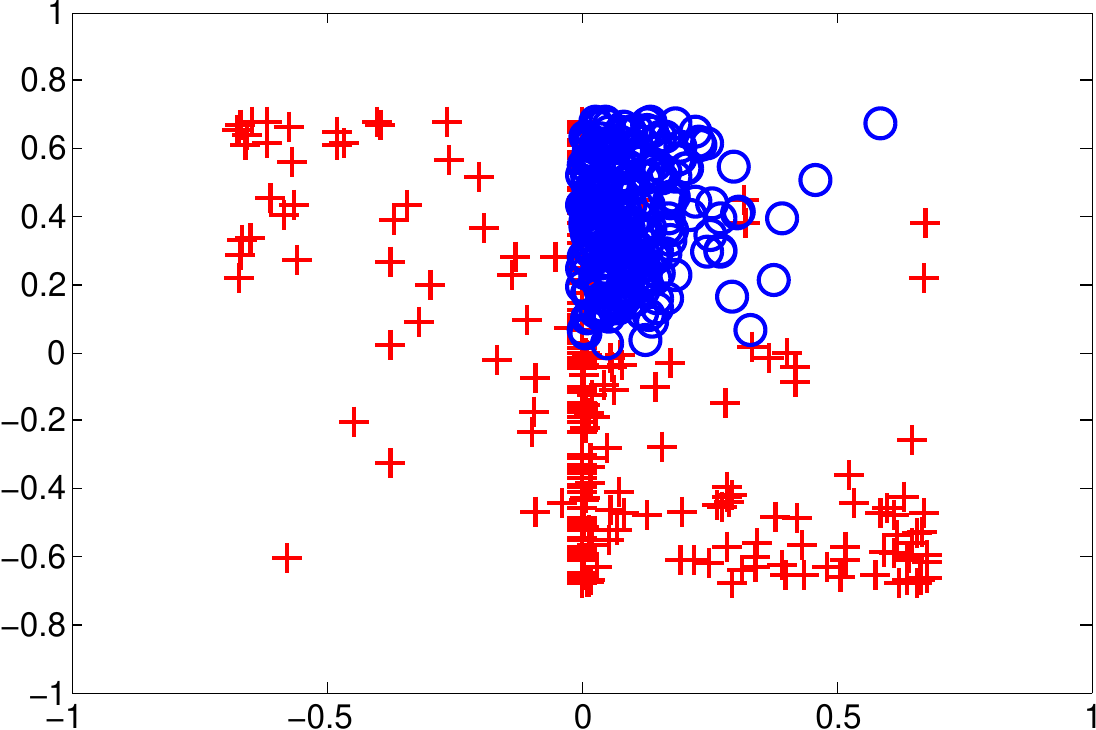}
                \label{fig:mapa}
        }
        \subfigure[Component wise exponential map]{
                \includegraphics[width=0.3\textwidth]{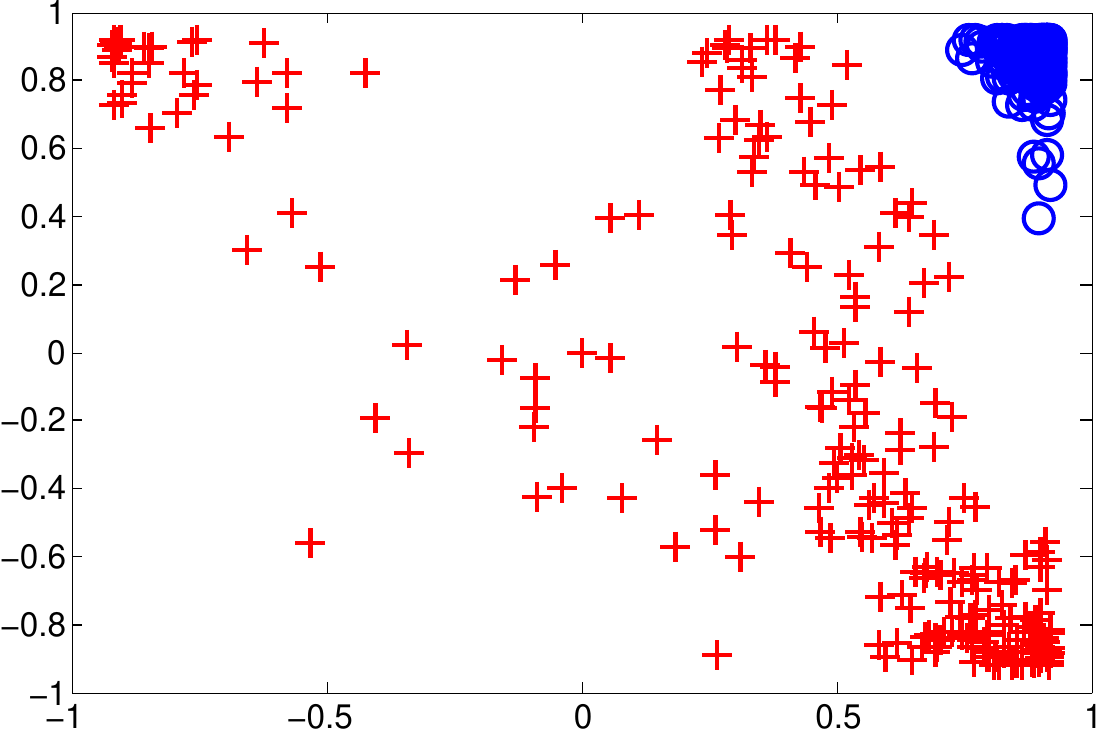}
                \label{fig:mapb}
        }
        \subfigure[Component wise linear map]{
                \includegraphics[width=0.3\textwidth]{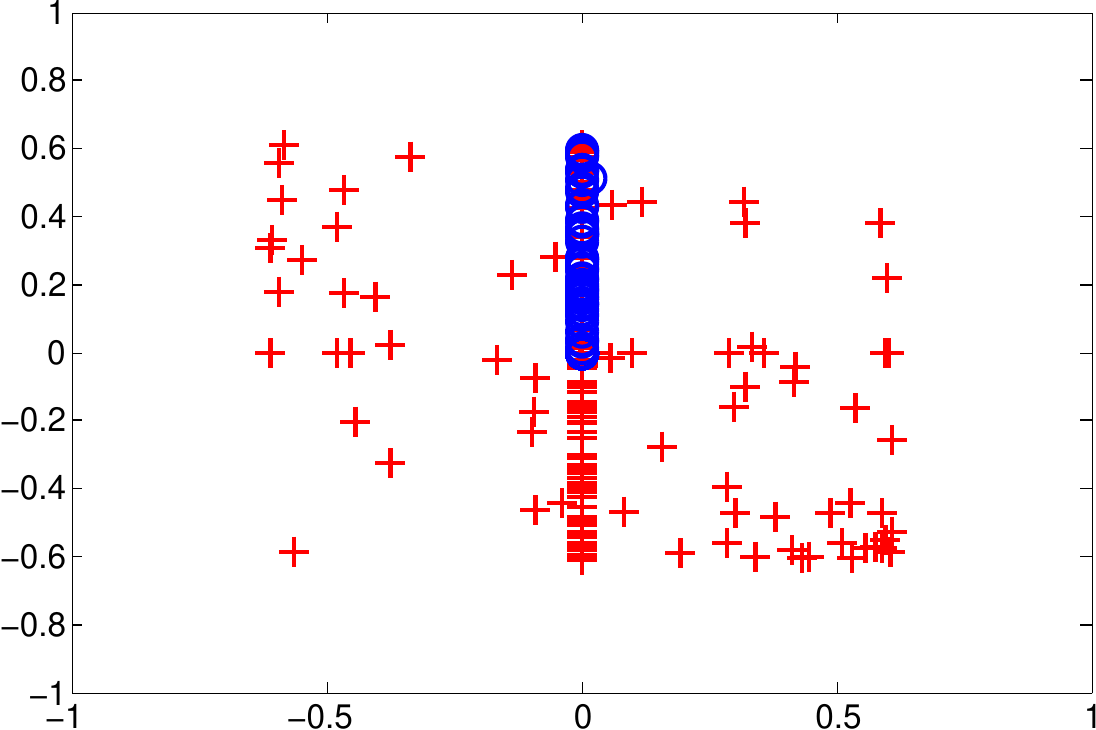}
                \label{fig:mapc}
        }
        \subfigure[Component wise squared map]{
                \includegraphics[width=0.3\textwidth]{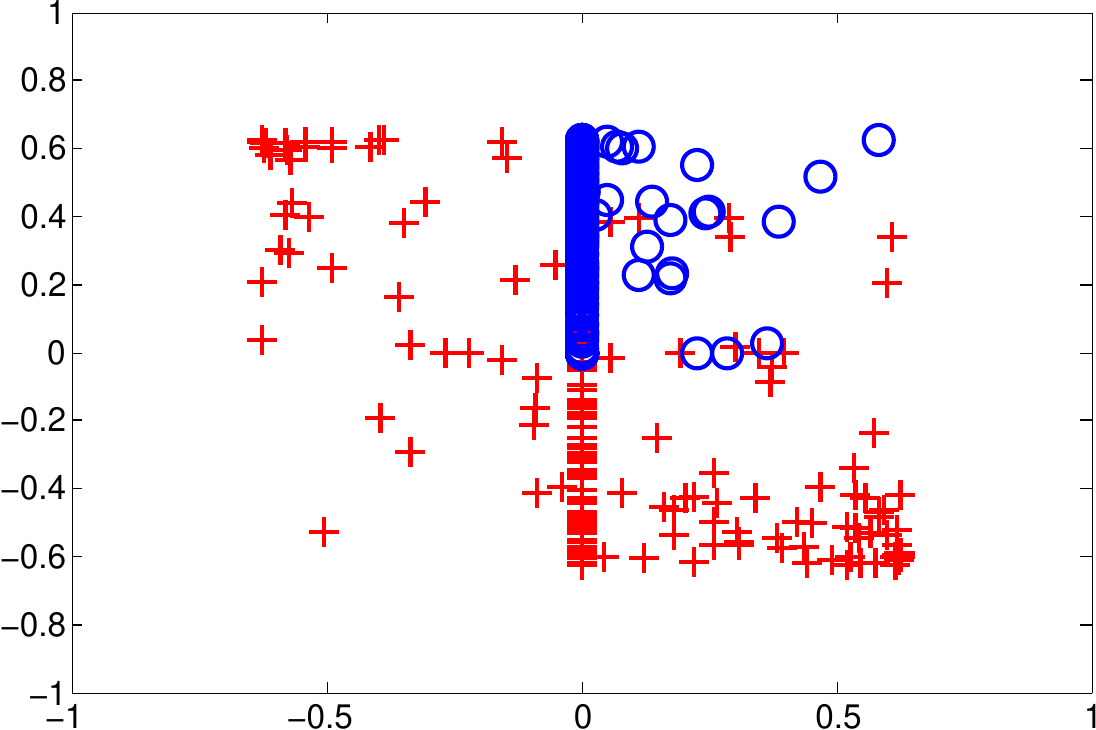}
                \label{fig:mapc}
        }
        \caption{Example of synthetic data and corresponding $\Xcal_c$ for different types of maps, with anchor point $a$.}\label{fig:map}
\end{figure*}

Let $\x_1, \x_2 \in \Rbb^d$ be two element of $\mathbb{R}^d$, and $\langle\x_1, \x_2\rangle = \x_1 ^\top \x_2$  the standard dot product .
Let the norm $\|\x\| = \sqrt{\langle\x, \x\rangle}$ and the metric $d(\x_1, \x_2) = \|\x_1 - \x_2\|$ be the standard (euclidean) norm and metric associated with $\langle\cdot,\cdot\rangle$.

We propose to explicitly build a Hilbert space $\Xcal_c$ as a subspace of the input space $\Rbb^d$ locally defined around an arbitrarily chosen center $\x_c \in \Rbb^d$. To enforce locality, we center the data on $\x_c$ and then apply a norm scaling function that tends to map the vectors to $0$ when the norm becomes to large. The mapping from $\Rbb^d$ to $\Xcal_c$ has the following expression:
\begin{align}
	\phi_c: & \Rbb^d \rightarrow \Xcal \\
	\nonumber & \x \mapsto h^{(c)}(\x)(\x - \x_c)
\end{align}

Where $h^{(c)}$ is a conformal map rendering the vicinity of the given center $\x_c$.
We show in Table~\ref{tab:map1} several examples of such mappings.
\begin{table}
\caption{Examples of conformal maps used in this work, with $|t|_+ = \max(0, t)$.}
\label{tab:map1}
\centering
	\begin{tabular}{c|c|c|c}
	name &  definition & bounded & smooth \\
	\hline
	$h_e^{(c)(\x)}$ &  $e^{-\gamma \| \x - \x_c \| }$ & - & - \\
	$h_g^{(c)}(\x)$ &  $e^{-\gamma \| \x - \x_c \|^2 }$ & - & $\checkmark$ \\
	$h_l^{(c)}(\x)$ &  $ \left| 1 -\gamma \| \x - \x_c \| \right|_+$ & $\checkmark$ & - \\
	$h_s^{(c)}(\x)$ &  $ \left| 1 -\gamma \| \x - \x_c \|^2 \right|_+$ & $\checkmark$ & $\checkmark$ 
  \end{tabular}
\end{table}
By definition, $\Xcal_c$ is a Hilbert space endowed with the following explicit dot product $k_c(\cdot, \cdot)$:
\begin{align*}
	k_c(\x_1, \x_2) =& \langle \phi_c(\x_1), \phi_c(\x_2) \rangle_{\Xcal_c}\\
\nonumber  =&	\left(h^{(c)}(\x_1) (\x_1-\x_c) \right)^\top \left( h^{(c)}(\x_2) (\x_2 -\x_c) \right)
\end{align*}
The properties of $\Xcal_c$ depends on the family of mappings $h$ chosen, which we consider to be a hyperparameter, and its subsequent parameters, namely $\{\gamma, \x_c\}$ in our examples.
We show in Figure~\ref{fig:map} examples of such mapping on synthetic data.
As we can see, many samples are mapped to or close to 0.
The effect is all the more visible using bounded maps, since samples outside of the support of $\phi_c$ are mapped exactly to 0.
However, the local geometry of the point cloud is preserved.

To allow for more sophisticated manipulations, we also consider component wise local mappings, in the form of:
\begin{align}
	\phi_c: & \Rbb^d \rightarrow \Xcal \\
	\nonumber & \x \mapsto \h^{(c)}(\x)\circ(\x - \x_c)
\end{align}
where $\circ$ is the entry wise (Hadamard) product.
With a slight abuse of notations, the mapping $\h: \Rbb^d \mapsto \Rbb^d$ now produces a vector rendering the vicinity of each component to the anchor point.
The explicit dot product corresponding to $\Xcal_c$ is then the following:
\begin{multline*}
	k_c(\x_1, \x_2)= \\ \left(h^{(c)}(\x_1)\circ (\x_1-\x_c) \right)^\top \left( h^{(c)}(\x_2)\circ (\x_2 - \x_c) \right)
\end{multline*}
Using the examples given in Table~\ref{tab:map1} on each component of input samples $\x$ produces the corresponding component wise mappings.
These mappings are also illustrated in Figure~\ref{fig:map}.

%

The main difference between the global and the component wise mappings is that the latter makes the assumption the components can be considered independently, which is the case if the features have been decorrelated (\textit{e.g.}, after a Karhunen-Lo\`eve transform).

It is clear that using only one of such kernels (either global or component wise) has less discriminatory capabilities than the simple linear kernel.
Indeed, since many features are mapped to 0 independently of their class, it is very likely that the classification problem becomes non linearly separable after the transform.
We thus consider the case where several locally linear kernels are used and summed into a single kernel:
\begin{equation}
	k(\x_1, \x_2) = \sum_c k_c(\x_1, \x_2)
\end{equation}
The resulting space $\Xcal$ is the concatenation of the $\Xcal_c$.
Problems that were not linearly separable in $\Rbb^d$ (for example the synthetic data of Figure~\ref{fig:map}) can be in $\Xcal$, provided a sufficient number of locally linear kernels with the right range of locality are used.

To choose these kernels, we consider the training set $\Acal$ of samples $\{\x_i\}_i$ of $\Rbb^d$ to be used to train the classifier.
We propose to perform a linear combination consisting of a locally linear kernel for each sample:
\begin{equation}
	k(\x_1, \x_2) = \sum_{\x_c \in \Acal} \beta_c k_c(\x_1, \x_2)
\end{equation}
We are now left with the optimization of $\beta$ so as to discard irrelevant kernels from the combination.
The next section devises the resulting optimization problem and proposes an algorithm to solve it.

\section{Multiple Locally Linear Kernel Machines}

Let $\Acal = \{ (\x_i, y_i) \}_{1 \leq i \leq n}$ be a training set of $n$ training samples $\x_i \in \Rbb^d$ and their associated labels $y_i \in \{-1, 1\}$.
The Multiple Locally Linear Kernel Machine (MLLKM) is then the optimal predictor according to the following primal problem:
\begin{align}
\label{eq:prim}  &\min_{w,\beta,\zeta} & \frac{1}{2}\sum_{m=1}^n \| w_m \|_{\mathcal{H}_m}^2 + C\sum_{i=1}^n \zeta_i\\
\nonumber & s.t. & \forall i, y_i \sum_m \sqrt{\beta_m} w_m^\top \phi_m(\x_i) \geq 1 - \zeta_i\\
\nonumber & & \forall i, \zeta_i \geq 0 \\
\nonumber & &\forall m, \beta_m \geq 0 \\
\nonumber & & \sum_m \beta_m = 1
\end{align}
Remark this is corresponds to a standard $\ell_1$-constraint MKL problem with the slight difference that the number of kernels is the same as the number of training samples.
On many occasions, $\ell_p$-MKL with $p>1$ are known to perform better than $\ell_1$ constraint~\cite{kloft09nips}, however the number of kernels in MLLKM is too high to allow for non sparse combinations.
In such case, the computational benefit of having a locally linear classifier would be lost to the very high number of linear predictors in the resulting combination.

To recover classical formulation of $\ell_1$-MKL, we consider the following Lagrangian of the primal problem with $\beta$ left in the constraints:
\begin{align*}
\nonumber	\mathcal{L}(w,\beta,\zeta,\alpha,\lambda) &=\frac{1}{2}\sum_m \| w_m \| ^2 + \sum_i \left(C  - \lambda_i - \alpha_i\right) \zeta_i \\
\nonumber	& - \sum_i \alpha_i\left[y_i \sum_m \sqrt{\beta_m} w_m^\top \phi_m(\x_i) - 1 \right]\\
	 s.t. & \quad  \forall m, \beta_m \geq 0 \\
  &\quad \sum_m \beta_m = 1
\end{align*}
The KKT stationary condition on $w$ states that:
\begin{align*}
	\frac{\partial \mathcal{L}}{\partial w_m^\star} = 0 = w_m^\star - \sum_i\alpha_i y_i \sqrt{\beta_m^\star}\phi_m(\x_i)
\end{align*}
Or equivalently:
\begin{align*}
	w_m^\star = \sum_i \alpha_i y_i \sqrt{\beta_m^\star} \phi_m(\x_i)
\end{align*}
Similarly, the stationary condition on $\zeta$ states that:
\begin{align*}
	\frac{\partial \mathcal{L}}{\partial \zeta_i} = 0 = C - \alpha_i - \lambda_i
\end{align*}
Since $\alpha$ and $\lambda$ are dual variables, the dual feasibility condition imposes their positiveness, and in particular implies a box constraint on $\alpha$:
\begin{align*}
\forall i, 	0 \leq \alpha_i \leq C
\end{align*}
A dual formulation of problem (\ref{eq:prim}) for $w$ and $\zeta$ variables is then obtained by injecting these conditions into the original formulation:
\begin{align*}
 \mathcal{D}(\alpha,\beta) &=\inf_{w,\zeta} \mathcal{L}(w,\beta,\zeta,\alpha,\lambda)\\
\nonumber & = \sum_i \alpha_i - \frac{1}{2}\sum_{i,j}\alpha_i \alpha_j y_i y_j \sum_m \beta_m k_m(\x_i,\x_j)\\
\nonumber	s.t.& \quad   \forall m, \beta_m \geq 0 \\
\nonumber   & \quad \sum_m \beta_m = 1
\end{align*}
This dual expression allows us to recover to the well known $\min\max$ formulation used for example in~\cite{rakoto08jmlr}:
\begin{align}
\label{prob:minmax}	\min_\beta \max_\alpha &\sum_i \alpha_i - \frac{1}{2} \sum_{i,j}\alpha_i \alpha_j y_i y_j \sum_m \beta_m k_m(\x_i, \x_j) \\
\nonumber	s.t. &\quad \forall i, 0 \leq \alpha_i \leq C \\
\nonumber	&\quad \forall m , \beta_m \geq 0\\
\nonumber	&\quad \sum_m \beta_m = 1
\end{align}
Since the number of kernels is equal to the number of training samples, large datasets lead to optimization problem intractable for current algorithms. In particular, most algorithms suppose all kernel matrices can be computed and stored in memory beforehand, which is clearly not the case when considering even medium sized datasets (\textit{i.e.}, more than 10k samples).
Consequently, we present a new algorithm called \emph{SequentialMKL}, for solving large MKL problems with reduced memory footprint that is suitable for training MLLKM.

\subsection{Sequential MKL}
\label{sec:smkl}
The main idea in SequentialMKL is to consider a reduced active set of kernels, solve the MKL problem for this reduced set, and then probe new kernels for inclusion in the set.
Solving the MKL problem with a reduced set of active kernels can be done efficiently using existing solver like~\cite{sonnenburg06jmlr,chapelle08nipsw} or~\cite{rakoto08jmlr}. In our case, we choose the reduced gradient approach of SimpleMKL of~\cite{rakoto08jmlr} since it is closely related to the inclusion criterion of new kernels. For the internal SVM solver, we use a variant of SDCA~\cite{shalevshwartz13jmlr} presented in Algorithm~\ref{alg:sdca}. We keep track of the outputs $\hat{\y}$ of the classifier and check for an early bail out criterion to improve the cost of each iteration.
\begin{algorithm}[t]
\caption{SDCA for solving SVM}
\begin{algorithmic}
\Function{SDCA}{$\y$, $K$, $E$, $C$}
\State $\hat{\y} \gets \bf 0$, $\alpha \gets \bf 0$, $e \gets 0$
\Repeat
	\State $L \gets$ random permutation of $\{1, ..., n\}$
	\For{$i \in L$}
		\State $g \gets 1 - \y_i \hat{\y}_i$
		\If{$g = 0$ \bf or ($g > 0$ \bf and $\alpha_i = C$) 
		\\ \hfill \bf or ($g < 0$ \bf and $ \alpha_i = 0$) \\\hspace{\algorithmicindent}\hspace{\algorithmicindent}\hspace{\algorithmicindent}}
		  \State skip $i$
		\EndIf
		\State $\alpha_{new} \gets \max(0, \min(\alpha_i + g/K_{ii}, C))$
		\State $\delta \gets \alpha_{new} - \alpha_i$
		\For{$j \in \{i, ..., n\}$}
		  \State $\hat{\y}_j \gets \hat{\y}_j + \delta \y_j K_{ij}$
		\EndFor		
	\EndFor
	\State $e \gets e + 1$
\Until $e \geq E$
\State\Return{$\alpha$}
\EndFunction
\end{algorithmic}
\label{alg:sdca}
\end{algorithm}
To design the criterion for inserting new kernels to the active set, we consider a Lagrangian of the $\min\max$ problem~\ref{prob:minmax} where the constraints of $\beta$ are taken into account:
\begin{multline*}
\nonumber	\mathcal{M}(\alpha,\beta,\mu, \nu) = 	\sum_i \alpha_i  - \sum_m \mu_m \beta_m - \nu\left(\sum_m\beta_m -1\right)\\ - \frac{1}{2} \sum_{i,j}\alpha_i \alpha_j y_i y_j \sum_m \beta_m k_m(\x_i, \x_j)
\end{multline*}
Now, the stationary condition at optimum $\beta^\star$ imposes that
\begin{align*}
	\nu^\star = -\mu_m^\star -\frac{1}{2}\sum_{i,j}\alpha_i^\star \alpha_j^\star y_i y_j k_m(\x_i, \x_j)
\end{align*}
Or, equivalently $\forall m, \forall l$:
\begin{multline*}
	\mu_m^\star + \frac{1}{2}\sum_{i,j}\alpha_i^\star \alpha_j^\star y_i y_j k_m(\x_i, \x_j) =  \\
	\mu_l^\star + \frac{1}{2}\sum_{i,j}\alpha_i^\star \alpha_j^\star y_i y_jk_l(\x_i, \x_j)
\end{multline*}
Let us denote $k_k$ a kernel with non-zero weight $\beta_k^\star$ at optimum, then $\mu_k^\star = 0$ by complementary slackness.
In turns, it means that $	\forall l \neq k$, 
\begin{align}
\nonumber \frac{1}{2}\sum_{i,j}\alpha_i^\star \alpha_j^\star y_i y_j k_k(\x_i, \x_j) =&
\label{eq:kcons}	\mu_l + \frac{1}{2}\sum_{i,j}\alpha_i^\star \alpha_j^\star y_i y_jk_l(\x_i, \x_j) \\
	\geq&\frac{1}{2}\sum_{i,j}\alpha_i^\star \alpha_j^\star y_i y_jk_l(\x_i, \x_j)
\end{align}
since by dual feasibility all $\mu_l^\star \geq 0$.
The criterion thus consists in finding a kernel of the open kernel set violating constraint (\ref{eq:kcons}) and adding it to the current active kernel set. 
For numerical stability reasons, we compute the largest $\frac{1}{2}\sum_{i,j}\alpha_i \alpha_j y_i y_j k_m(\x_i, \x_j)$ among non zero weight kernels $k_m$.
Remark that kernels respecting Equation~(\ref{eq:kcons}) have zero weight and $\mu_m$ equals to the difference of gradients with $k_k$, meaning the KKT conditions are respected.

The full SequentialMKL algorithm is presented in Algorithm~\ref{alg:smkl}.
We alternate between optimizing the kernel weights and search for new kernel to insert in the active set, until no new kernel can be inserted.
\begin{figure*}[!t]
	    \centering
        \subfigure[Gaussian map]{
                \includegraphics[width=0.22\textwidth]{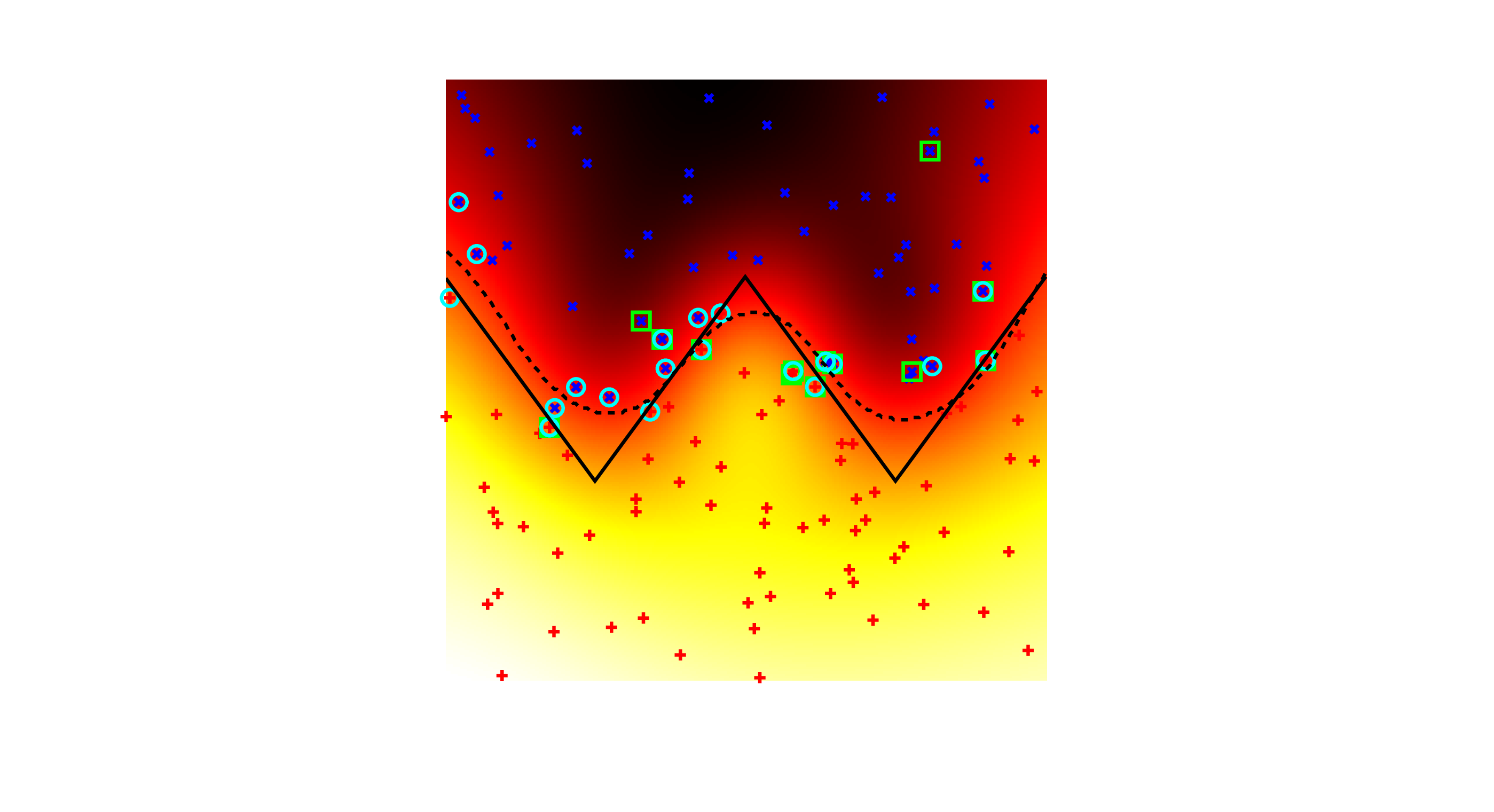}
                \label{fig:ori}
        }
        \subfigure[Square map]{
                \includegraphics[width=0.22\textwidth]{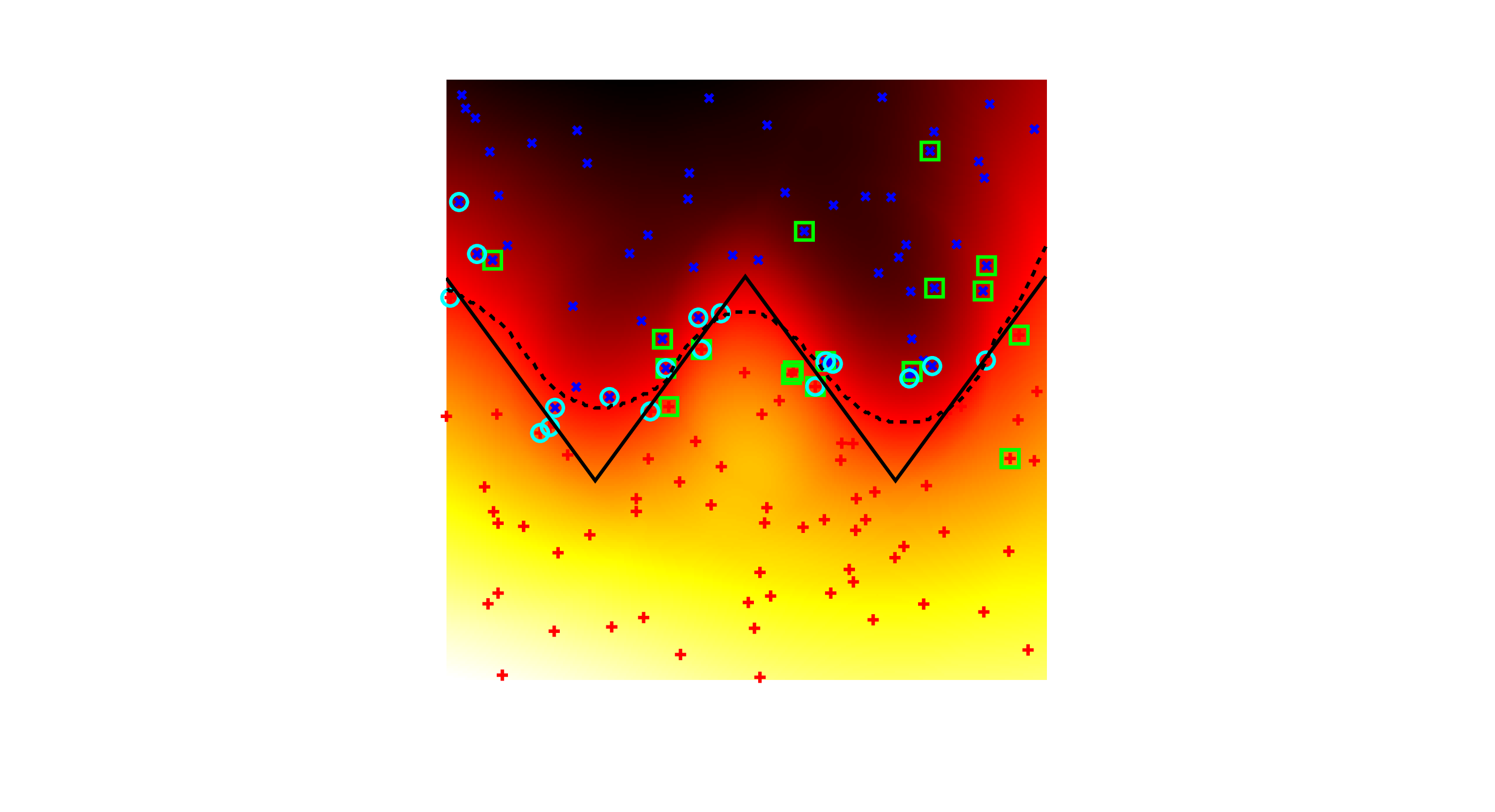}
                \label{fig:mapa}
        }
        \subfigure[Component Gaussian map]{
                \includegraphics[width=0.22\textwidth]{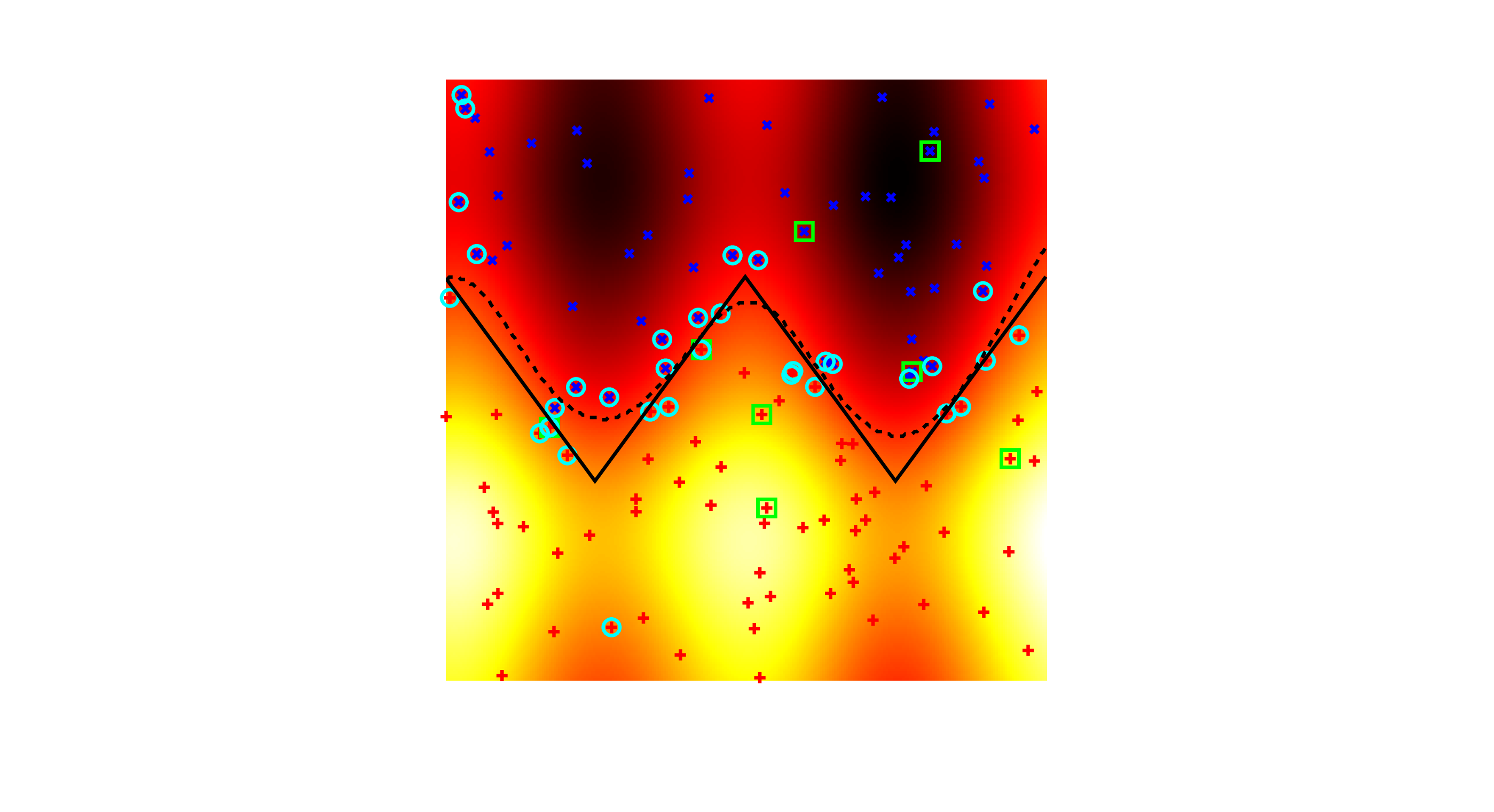}
                \label{fig:ori}
        }
        \subfigure[Component Square map]{
                \includegraphics[width=0.22\textwidth]{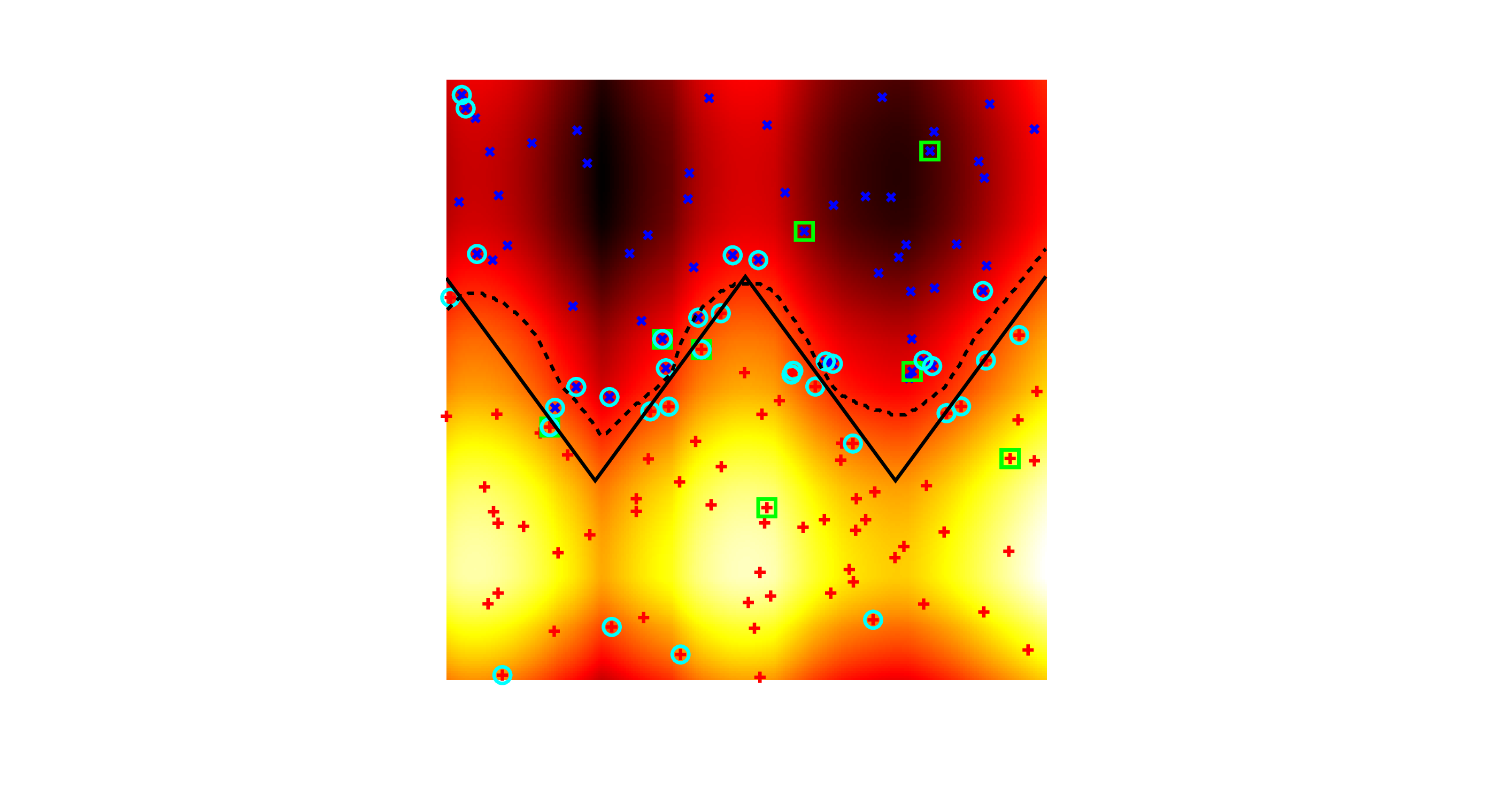}
                \label{fig:mapa}
        }
        \caption{Example of synthetic data classification task. The true boundary is shown in plain black line, while the classifier boundary is in dashed black line. The circles denote the support vectors, while the squares denote the support kernels.}\label{fig:toys}
\end{figure*}
\begin{algorithm}[t]
\caption{SequentialMKL}
\begin{algorithmic}
\Function{train}{$\{k_m\}$, $\x$, $\y$, $E$, $C$}
\State $\Scal \gets k_1$, $\Ocal \gets \{k_m\} \backslash k_1$
\State $\beta \gets \bf 1$, $\alpha \gets \bf 0$
\Repeat
  \State $K \gets \sum_{k_m \in \Scal} \beta_m k_m$
  \State $\alpha \gets$ SDCA($\y$, $K$, $E$, $C$)
  \State $term \gets \bf true$
  \State $g \gets \max_{k_m \in \Scal} \frac{1}{2}\sum_{i,j}\alpha_i\alpha_j y_i y_j k_m(i,j)$
  \For{$k_m \in \Ocal$}
    \If{$\frac{1}{2}\sum_{i,j}\alpha_i\alpha_j y_i y_j k_m(i,j) > g$}
      \State $term \gets \bf false$
      \State $\Scal \gets \Scal + k_m$, $\Ocal \gets \Ocal \backslash k_m$
    \EndIf
  \EndFor
  \State $\beta \gets$ SolveMKL($\Scal$, $\beta$, $\x$, $\y$, $E$, $C$)
  \For{$\beta_m \in \beta$}
    \If{$\beta_m = 0$}
      \State $\Scal \gets \Scal \backslash k_m$, $\Ocal \gets \Ocal + k_m$
    \EndIf
  \EndFor
\Until $term$
\State\Return{$\alpha$, $\beta$}
\EndFunction
\end{algorithmic}
\label{alg:smkl}
\end{algorithm}
In practice, we found that inserting more than one kernel at a time do not increase the time taken to update the weights, while significantly reducing the number of iterations of the outer loop.
Indeed, there is a tradeoff between inserting many kernels and performing several passes over the entire kernel set that can be roughly decided by the maximum number of Gram matrices that can be cached in memory to allow fast training in the inner loop.

Moreover, SequentialMKL can very simply be extended to an online variant where the kernels are streamed.
In such case, each time a new kernel is considered, it is probed for insertion using the gradient criterion, and all weights are consequently re-optimized (both $\alpha$ and $\beta$).
To obtain the true MKL solution, all previously discarded kernels have to be probed after each new insertion, which can be done using a reprocess function akin to the one in LaSVM~\cite{bordes05jmlr}.
When working on a budget, discarded kernels can be forgotten, but then the solution given by SequentialMKL is only approximated unless many passes over the whole kernel set are made.
In the specific case of MLLKM, a fully online algorithm can be efficiently designed, where training samples are streamed and the corresponding kernels are probed at the same time.

\subsection{Inference computational cost}
Like in many MKL problem, the trained classification function is the following:
\begin{equation}
	f(x) = \sum_i \alpha_i \sum_c \beta_c k_c(x_i, x).
\end{equation}
For the quasi-conformal mapping, the expression of $f$ can be further developed using the explicit feature space:
\begin{equation}
	f(x) = \sum_i \alpha_i \sum_c \beta_c h_c(x_i)h_c(x)(x_i-x_c)^\top(x-x_c).
\end{equation}
Most of its computation can then be accelerated using this explicit mapping:
\begin{equation}
	f(x) = \sum_c h_c(x)(x-x_c)^\top w_c, 
\end{equation}
with
\begin{equation}
	w_c = \beta_c \sum_i \alpha_i h_c(x_i)(x_i - x_c).
\end{equation}
The same can be done for the component wise mappings.

Using this form, the inference cost is expected to be low since the algorithm is set to have a sparsity pattern on $\beta$ due to the $\ell_1$ norm constraint.
Moreover, the evaluation in this form independent of the number of support vectors, just as for linear SVM.
Remark that this inference cost is valid both in terms of number of operations and in memory requirement.
In case where the classifier has to fit a low memory hardware at inference time, MLLKM has clearly the advantage that only a few number of parameters have to be stored compared with kernel SVM trained on a huge training set.

\subsection{Parameter selection}

For the mapping function of Table~\ref{tab:map1}, the $\gamma$ parameters have to be tuned to take into account the scale of the data.
To avoid the burden of doing so by cross-validation, a sampling over a reasonable segment (\textit{e.g.}, $[0.01, 10]$) can be made and all the corresponding locally linear kernels related to the training samples are considered in the initial kernel set.
Although it raises dramatically the total number of kernels, we found this solution to be more efficient in training time than choosing the right parameter using cross-validation.
Furthermore, it allows for different parameters for different kernels, which betters renders the different scales of the data.

\section{Experiments}
\label{sec:exp}
We test our algorithms on 2 different datasets.
The first dataset is a synthetic binary classification task with highly non-linear boundary between the classes.
Then we compare our algorithm performances on well known datasets of the UCI repository.

\subsection{Synthetic data}

We first present experiments on a synthetic dataset containing two uniformly distributed classes, separated by a piece-wise linear boundary.
Figure~\ref{fig:toys} shows the output values of the classifier as a heat map, the true boundary in continuous line, the boundary of the classifier in dashed line, support vectors in circles and active kernels in squares. $C$ was set to 1000.

As we can see, we can achieve a curved boundary thanks to the selected locally linear kernels.
In particular, the square and component-wise square maps have higher curvature (infinite in the case of component wise mapping), which may be due to the bounded support of such maps.
The component wise maps have less active kernels, although they provide a comparable boundary.
Remark the support vectors are only shown to measure the extend of the support domain for the boundary, since they are not used in the computation of the output. 
Using conformal map leads to support vectors concentrated around the boundary, while using component siwe maps leads to support vectors scattered inside the maximum $\ell_\infty$ bounding box the the active kernel set.

\begin{table*}[t!]
\caption{Results on several UCI datasets. Methods are grouped into categories representing MLLKM with varying kernel type, linear and locally linear methods and non-linear methods.}
\label{tab:res}
\centering
\scshape
\begin{tabular}{|l|c|c|c|}
\hline
method & accuracy & inference time (ms) & \#kernels$\times$ \#support vectors \\
\hline
\hline
\multicolumn{4}{|c|}{ionosphere}\\
\hline
MLLKM (Gaussian) & \textbf{ 94.0} $\pm$  2.3	&	  \textbf{1.0} $\pm$  0.0	&	 22.5 $\pm$  7.9 $\times$ 1  \\
\hline
MLLKM (Square) & 92.9 $\pm$  2.8	&	  2.4 $\pm$  3.0	&	 25.9 $\pm$  9.0 $\times$ 1  \\
\hline
MLLKM (Component Gaussian) &  91.7 $\pm$  1.1	&	  8.0 $\pm$  2.2	&	 27.0 $\pm$  5.7 $\times$ 1 \\
\hline
MLLKM (Component Square) & 91.4 $\pm$  2.4	&	  1.9 $\pm$  0.8	&	 25.3 $\pm$  7.4 $\times$ 1  \\
\hline
\hline
SAG &  79.1 $\pm$  4.3	&	  0.0 $\pm$  0.0 & 1 $\times$ 1  \\
\hline
LLSVM & \textbf{87.3} $\pm$  2.2	&	  \textbf{1.5} $\pm$  1.4  & 32 $\times$ 1 \\
\hline
\hline
LaSVM (Gaussian) &   94.3 $\pm$  2.2	&	  1.2 $\pm$  0.6  &  1  $\times$ 53.3 $\pm$  4.3  \\
\hline
SimpleMKL & \textbf{95.1} $\pm$  2.2	&	 \textbf{18.9} $\pm$  8.1	&	  9.5 $\pm$  4.4	$\times$	 159.7 $\pm$ 13.1 \\
\hline
SequentialMKL & 93.3 $\pm$  3.1	&	 55.2 $\pm$ 22.5	&	 15.9 $\pm$  8.1	$\times$	 141.8 $\pm$ 30.9 \\
\hline
\hline
\multicolumn{4}{|c|}{sonar}\\
\hline
MLLKM (Gaussian) &  \textbf{81.0} $\pm$  2.8	&	   \textbf{0.6} $\pm$  0.5	&	 17.6 $\pm$  5.5 $\times$ 1  \\
\hline
MLLKM (Square) & 80.2 $\pm$  3.5	&	  0.8 $\pm$  0.4	&	 18.4 $\pm$  5.4 $\times$ 1  \\
\hline
MLLKM (Component Gaussian) &  77.8 $\pm$  2.4	&	  4.9 $\pm$  1.8	&	 19.8 $\pm$  7.3 $\times$ 1 \\
\hline
MLLKM (Component Square) & 80.2 $\pm$  3.0	&	  1.0 $\pm$  0.4	&	 23.1 $\pm$  5.0 $\times$ 1  \\
\hline
\hline
SAG &  62.5 $\pm$  5.9	&	  0.0 $\pm$  0.0 & 1 $\times$ 1  \\
\hline
LLSVM & \textbf{67.8} $\pm$  5.4	&	  \textbf{0.5} $\pm$  0.5  & 32 $\times$ 1 \\
\hline
\hline
LaSVM (Gaussian) &    79.0 $\pm$  3.5	&	  0.5 $\pm$  0.5   &  1  $\times$ 73.6 $\pm$  4.1 \\
\hline
SimpleMKL & 84.6 $\pm$  4.4	&	 10.7 $\pm$  1.9	&	  8.5 $\pm$  1.1	$\times$	 138.1 $\pm$  2.9 \\
\hline
SequentialMKL & \textbf{86.3} $\pm$  2.7	&	 \textbf{15.4} $\pm$ 12.0	&	 11.2 $\pm$  9.2	$\times$	 124.5 $\pm$ 26.2\\
\hline
\hline
\multicolumn{4}{|c|}{heart}\\
\hline
MLLKM (Gaussian) &  81.6 $\pm$  3.0	&	  0.6 $\pm$  0.5	&	 26.0 $\pm$  8.3 $\times$ 1  \\
\hline
MLLKM (Square) & 81.4 $\pm$  4.1	&	  0.5 $\pm$  0.5	&	 27.1 $\pm$ 11.6 $\times$ 1  \\
\hline
MLLKM (Component Gaussian) &  \textbf{82.2} $\pm$  4.9	&	  \textbf{2.2} $\pm$  0.4	&	 27.3 $\pm$  6.3 $\times$ 1 \\
\hline
MLLKM (Component Square) & 81.1 $\pm$  2.8	&	  0.8 $\pm$  0.6	&	 26.8 $\pm$  6.9 $\times$ 1  \\
\hline
\hline
SAG &  \textbf{83.0} $\pm$  3.5	&	  \textbf{0.1} $\pm$  0.3 & 1 $\times$ 1  \\
\hline
LLSVM & 80.4 $\pm$  4.1	&	  0.3 $\pm$  0.5  & 32 $\times$ 1 \\
\hline
\hline
LaSVM (Gaussian) &   79.6 $\pm$  2.4	&	  0.5 $\pm$  0.5   &  1  $\times$ 74.7 $\pm$  5.2 \\
\hline
SimpleMKL & \textbf{80.2} $\pm$  2.9	&	 \textbf{10.6} $\pm$  2.2	&	  7.4 $\pm$  0.9	$\times$	 163.2 $\pm$  5.7 \\
\hline
SequentialMKL & 80.0 $\pm$  5.3	&	 12.1 $\pm$  2.5	&	  8.0 $\pm$  1.9	$\times$	 164.9 $\pm$ 19.0 \\
\hline
\hline
\multicolumn{4}{|c|}{diabetes}\\
\hline
MLLKM (Gaussian) & 75.3 $\pm$  3.2	&	  0.9 $\pm$  0.5	&	 24.0 $\pm$  5.4	$\times$	 1	\\
\hline
MLLKM (Square) &	 74.9 $\pm$  3.6	&	  0.8 $\pm$  0.6	&	 28.1 $\pm$ 13.0	$\times$	 1	\\
\hline
MLLKM (Component Gaussian) &	 \textbf{76.0} $\pm$  4.1	&	  \textbf{2.0} $\pm$  0.9	&	 29.6 $\pm$  7.3	$\times$	 1 \\
\hline
MLLKM (Component Square) & 72.6 $\pm$  3.5	&	  1.1 $\pm$  1.1	&	 38.6 $\pm$ 11.3	$\times$	 1	\\
\hline
\hline
SAG	& 66.1 $\pm$  9.0	&	  0.0 $\pm$  0.0	&	  1 $\times$ 1	\\
\hline
LLSVM &	 \textbf{74.2} $\pm$  2.7	&	  \textbf{2.3} $\pm$  2.0	&	  32	$\times$	  1	\\
\hline
\hline
LaSVM (Gaussian) &	 \textbf{77.1} $\pm$  3.2	&	  \textbf{3.6} $\pm$  1.5	&	  1 $\times$ 141.9 $\pm$  7.4	\\
\hline
SequentialMKL &	 69.5 $\pm$  2.8	&	 55.0 $\pm$ 118.6	&	  7.1 $\pm$  9.7	$\times$	 216.6 $\pm$ 21.7	\\
\hline
SimpleMKL &	 69.5 $\pm$  4.1	&	  8.9 $\pm$  1.0	&	  3.3 $\pm$  0.5	$\times$	 223.5 $\pm$  4.5	\\
\hline
\end{tabular}
\end{table*}
\subsection{UCI dataset}
We evaluated our MLLKM on several UCI datasets (ionoshpere, sonar, heart and diabetes), following the procedure of~\cite{rakoto08jmlr} which is the most used in MKL evaluations.
For fairness of evaluation with respect to the computational time, we implemented both MLLKM and SequentialMKL using the JKernelMachines library~\cite{picard13jmlr} which already contains several algorithms against which we compared our work, \textit{i.e.}, LaSVM~\cite{bordes05jmlr}, SAG~\cite{leroux12nips}, LLSVM~\cite{ladicky11icml} and SimpleMKL~\cite{rakoto08jmlr}.
We performed 10 random split of the data where $70\%$ was kept for training, and measured the average accuracy.
$C$ was arbitrarily set to 100 for all algorithms and all datasets.

When using MKL, we used the standard setup that consists in adding Gaussian kernels for each component (10 bandwidths) as well as homogeneous and inhomogeneous polynomial kernels of degrees up to 3. The total number of kernels in such case is 19 times the dimension of the input space.
For LLSVM, we arbitrarily set the number of anchor points to 32, while the bandwith of the Gaussian kernel for LaSVM was set using crossvalidation.

We show in Table~\ref{tab:res} the accuracy, the mean inference time and the number of active kernels and support vectors for all methods.
The results are grouped into 3 categories: Our method with varying kernel types, linear or locally linear methods and non-linear methods.
As we can see MLLKM is in general close to non-linear methods in terms of accuracy, while having a much reduced inference time.
Except for the heart dataset, MLLKM always outperforms linear methods by a fair margin while having a comparable inference time.

One interesting property of MLLKM is with respect to the storage cost of the classifier, which is proportional to the number of selected kernels.
On the contrary, Kernel SVM and MKL have to store the support vectors.
As we can see, the number of selected kernels in MLLKM is much lower than the number of support vector in Kernel SVM and MKL (by a factor of 5 to 10 in case of MKL), which means the resulting classifier is easier to embed in a low memory device even when trained on a large dataset.

\section{Conclusion}
In this paper, we presented a new classifier named Multiple Locally Linear Kernel Machine based on the combination of locally linear kernel.
The proposed approach has several advantages: First it fits perfectly between high accuracy kernel SVM that have a high inference cost and lower accuracy linear SVM with a low inference cost.
Second, the proposed problem is equivalent to $\ell_1$-MKL, which give some guarantees the training procedure always gives a good solution.

To train the MLLKM, we proposed a generic $\ell_1$-MKL algorithm that is able to cope with a very high number of kernels.
The procedure is based on a reduced active set of kernels and scans the remaining kernels for insertion.
As such, 2 extensions can easily be considered: An online kernels MKL where kernels are streamed while the training set is known, and a full online where both training samples and kernels are streamed.
This later case is the one of an online MLLKM since the kernels are build around the training samples.

Finally, our experiments show that MLLKM is able to obtain accuracies comparable to non-linear classifier such as MKL, while having an inference time comparable to linear methods.

\bibliographystyle{icml2015}
\bibliography{picard_mllkm}

\appendix
\section{ICML'15 reviews}
In this section, we show the reviews received at ICML'15 where the paper was ultimately rejected.

\subsection{Reviewer 3}
\textbf{Overall Rating:}	Strong reject\\
\textbf{Reviewer confidence:}	Reviewer is an expert\\
\textbf{Detailed comments for the authors}\\
\textit{Summary}
-------
The paper looks at the problem of accelerated prediction times with non-linear classifiers. To this end the paper proposes a multiple kernel learning (MKL) approach. The base kernels chosen are "locally linear" - to understand these local kernels better, let me take an extreme setting. Each local kernel is centered around an "anchor point" $c$. For points $x,y$ within a radius $r$ of $c$, the linear kernel is preserved $k_c(x,y) = (x-c)'(y-c)$. If either point $x,y$ is outside this radius, $k_c(x,y) = 0$.

Of course the paper uses smooth variations using conformal maps to implement these kernels which reweigh the local kernel according to the distance of the points from the anchor point. Radial as well as coordinate-wise maps are considered as well. Every training point is considered a potential anchor point and a sparse combination of these kernels is learnt using the SequentialMKL approach which alternates between updating an active set of kernels and choosing an appropriate combination of these active kernels.

\textit{Quality}
-------
The paper delves upon a slightly stale methodology of "locally linear" methods. These are intuitive methods that are easy to explain but hard to implement and even harder to scale to large problems.

\textit{Clarity}
-------
The paper is well written. Section 4 could do with proof checking to remove spelling/grammatical mistakes.

\textit{Originality}
-----------
The idea itself seems novel although it effectively promotes similar classification characteristics as other local methods. The novel part in the paper seems to be its truly non-parametric approach to locality wherein every training point is considered a potential anchor point. Most such works implement locality by clustering the data using k-means and then choosing the cluster centers as anchor points.

\textit{Significance}
------------
The approach seems not very promising practically as recent literature (please see references below) already has several examples of methods that outperform locally linear methods comprehensively in terms of accuracy, as well as prediction time.

\textit{Comments}
--------
- Since a primary goal of the paper seems to be reduced prediction time, recent works in that direction such as (Hsieh et al, 2014) and (Jose et al, 2013) need to be referenced, especially since these methods handily outperform locally linear methods.
- The experimental work is a bit underwhelming - all prediction times are in milliseconds. Datasets used are tiny at a time when other such papers are performing experiments on the MNIST 8Million dataset. Moreover the proposed method does not present a strong case for choosing it over other methods.
- Since this is primarily an experimental piece of work with little theoretical results, a much more comprehensive comparison to the state of the art would have to be performed to argue in favor of the proposed method.

\textbf{References}
----------

C.-J. Hsieh, S. Si, and I. S. Dhillon, Fast Prediction for Large-Scale Kernel Machines, NIPS 2014.

C. Jose, P. Goyal, P. Aggrwal and M. Varma. Local deep kernel learning for efficient non-linear SVM prediction, ICML 2013.

\textbf{[REQUIRED AFTER REBUTTAL]}\\
\emph{I have read and considered the authors' rebuttal.}	Yes\\
\emph{Post-rebuttal Comments.}	I would encourage the authors to support their method with comprehensive experiments on large scale datasets and comparisons to the state-of-the-art.

\subsection{Reviewer 4}
\textbf{Overall Rating:}	Weak reject\\
\textbf{Reviewer confidence:}	Reviewer is knowledgeable\\
\textbf{Detailed comments for the authors}	

In combining multiple locally linear kernels, the idea is to start with one candidate kernel centered on each training point and let regularization pick the ones that are necessary and remove the ones that are not; simiarly for the kernel spread ($\gamma$), a bunch of values are made available as candidates for each center and the regularizer keeps the necessary ones. This reads like a nice idea but ideally the positiion (anchor) can be different from any of the training points (that's the advantage of methods like LLSVM and LMKL) and similarly the ideal spread can be different from any of the predefined candidates. In a high dim problem with a small data set, limiting the anchors to the training data may be restrictive.

Concerning the experimentation: The synthetic problem is simple (and what can you say about the chosen $\gamma$?) and comparison is done on four small UCI data sets. The stdev vaues on the latter seem large with respect to differences between the means; are those differences statistically significant? (Why are some results printed in boldface?)

And do not start the abstract with "In this paper"--it is redundant.

\textbf{[REQUIRED AFTER REBUTTAL]}\\
\emph{I have read and considered the authors' rebuttal.}	Yes\\
\emph{Post-rebuttal Comments.}	More comparison with existing work is needed. My original review stands.

\subsection{Reviewer 6}
\textbf{Overall Rating:}	Weak reject\\
\textbf{Reviewer confidence:}	Reviewer is an expert\\
\textbf{Detailed comments for the authors}	

The goal of this paper is to improve traditional multiple kernel learning so that to fit the gap of the inference/prediction speed and accuracy between linear SVM and kernel SVM. There are three main contributions: (1) use l1 regularized multi-kernel learning objective aligned with locally linear kernels; (2) speed up the optimization by considering a reduced set of kernels and then probing new kernels after solving MKL in the reduced set; (3) new feature mapping. The experiments are conducted on synthetic and public datasets showing the efficiency of the proposed method. The paper is well written and organized.

I have several questions/suggestions for this paper:

1: It would be great to show the motivation for the new feature mapping in eq(1) or explain the advantage of this feature mapping over traditional kernels(Gaussian kernel or linear kernel). In the experimental part, it would be interesting to vary kernel functions in your MKL framework, e.g., replace feature mapping in eq(1) with linear feature mapping, etc., so that readers can believe that the proposed new feature mapping can benefit the classification accuracy.

2: I guess I am still confused why this method is memory efficient. Would all the kernel matrices be computed and stored in advance to perform kernels selection? I guess the total memory requirement is still $O(n^2)$. Maybe authors can provide more explanation over the memory requirement.

3: There are $x_c$ in the new feature mapping. Do you choose them by kmeans clustering? Or $x_c$ needs to be learned through optimization. Is $c=n$ (the number of data samples)? Will the prediction be faster if using kmeans centers as $x_c$?

4: Some comparisons with state-of-the-art fast inference algorithms are missing: for example LDKL in [1] and DC-Pred++ in [2]. Also as claimed in the paper, the method is scalable, and it would be interesting to show some experimental results on large-scale datasets. All the datasets tested in the paper have less than 1000 data points.

[1] C. Jose, P. Goyal, P. Aggrwal, and M. Varma, Local deep kernel learning for efficient non-linear svm prediction, in ICML, 2013.
[2] C.-J. Hsieh, S. Si, and I.S.Dhillon, Fast Prediction for Large-Scale Kernel Machines, in NIPS, 2014.

5: If I understood correctly, the prediction/inference time for the proposed algorithm is \# of kernel * number of support vectors in each kernel. So the fourth column of Table 2 should correspond to the third column of Table 2. For ionosphere dataset, the MLLKM (Gaussian) is with the similar \# of kernel * number of support vectors with MLLKM (component gaussian), however the former takes only 1/8 inference time of the latter one. So authors might want to clarify the difference.

\textbf{[REQUIRED AFTER REBUTTAL] }\\
\emph{I have read and considered the authors' rebuttal.}	No

\subsection{Meta-Reviewer 1}
\textbf{Overall Rating:}	Reject\\
\textbf{Detailed Comments}	
The reviewers are in agreement that this paper is not yet ready for publication. All three reviewers consistently felt, both before the reviewer discussion and afterwards, that the paper has major weaknesses that need to be addressed before the paper can be accepted for publication. Both Area Chairs agree with the reviewer assessment. The rebuttal is unconvincing in that DCPred++ should have been cited and discussed in the paper even if experimental comparisons could not have been carried out by the time of submission. Furthermore, some comparative results should have been presented in the rebuttal -- it should have been possible to run the proposed algorithm on the DCPred++ data sets even if DCPred++ could not have been implemented by the authors till then. Finally, comparisons should also have been performed to LDKL since both the code and data sets are publically available. The authors should take the reviewer feedback into account while preparing future versions of the manuscript.

\end{document}